\documentclass[twoside,leqno,twocolumn]{article}

\usepackage[letterpaper]{geometry}
\usepackage{stfloats}
\usepackage{siamproceedings}

\usepackage[T1]{fontenc}
\usepackage{amsfonts}
\usepackage{graphicx}
\usepackage{epstopdf}
\usepackage{enumitem}
\usepackage{booktabs}

\usepackage{algorithmicx}
\usepackage{algorithm}
\usepackage{algpseudocode}
\ifpdf
  \DeclareGraphicsExtensions{.eps,.pdf,.png,.jpg}
\else
  \DeclareGraphicsExtensions{.eps}
\fi

\newsiamremark{remark}{Remark}
\newsiamremark{hypothesis}{Hypothesis}
\crefname{hypothesis}{Hypothesis}{Hypotheses}
\newsiamthm{claim}{Claim}

\usepackage{amsopn}

\begin{document}

\newcommand\relatedversion{}

\title{\Large Adaptive Sampling for Automated Post-Disaster Rapid Damage Assessment via Level-Set Cost-Aware Bayesian Optimization}
    \author{Boyang Xu\thanks{School of Computing and Augmented Intelligence, Arizona State University (\email{boyangxu@asu.edu}, \email{haoyan@asu.edu}). Hao Yan is the corresponding author.}
\and Mostafa Reisi Gahrooei\thanks{Department of Industrial and Systems Engineering, University of Florida (\email{mreisigahrooei@ufl.edu}).}
\and Mohammad Ilbeigi\thanks{Department of Civil, Environmental and Ocean Engineering, Stevens Institute of Technology (\email{milbeigi@stevens.edu}).}
\and Hao Yan\footnotemark[1]}

\date{}

\maketitle


\fancyfoot[R]{\scriptsize{Copyright \textcopyright\ 2026 by SIAM\\
Unauthorized reproduction of this article is prohibited}}





\begin{abstract} Natural disasters frequently inflict severe damage to the built environment, which demands a rapid, reliable, and cost-effective damage assessment for emergency response. However, traditional methods for post-disaster damage assessment often rely on static, labor-intensive data collection strategies that can be prohibitively expensive and struggle to adapt to dynamic post-disaster conditions. In this study, we propose a cost-aware Bayesian optimization framework combined with level-set estimation that continuously guides autonomous data collectors, e.g., an unmanned aerial vehicle (UAV), toward the most informative regions. By dynamically updating damage estimates across different geographic zones, our approach systematically reduces uncertainty while minimizing operational costs. The proposed framework is first validated using a controlled synthetic toy study, demonstrating the agent's ability to efficiently trace damage boundaries, recover the underlying damage map, and rapidly reduce predictive uncertainty. Furthermore, the approach is evaluated using high-fidelity disaster data generated by the Regional Resilience Determination (R2D) software. The results of the algorithm provide accurate and timely damage estimates that support informative and fast emergency response.
\end{abstract}

\section{Introduction.}
Natural disasters such as hurricanes, earthquakes, floods, and wildfires often cause extensive damage to critical infrastructure and the broader urban environment, disrupting essential services and endangering human lives. Evaluation of disaster impacts on the physical and operational status of buildings and infrastructure systems is essential for enabling timely, informed decision-making during emergency response efforts and serves as a cornerstone for cost analysis, strategic planning, and coordination~\cite{lozano2023data}. Post-disaster damage assessment (PDDA) plays a pivotal role in this process by providing necessary information on the extent and severity of damage. When a natural disaster strikes a built environment, three types of PDDA are conducted to determine the impact and magnitude of the disaster: (1) Rapid Damage Assessment (RDA); (2) Preliminary Damage Assessment (PDA); and (3) Substantial Damage Assessment (SDA) ~\cite{hodde2012damage}. These damage assessment processes overlap and create a continuous process that begins immediately after a disaster and continues into and beyond the post-impact period. However, their focus, required accuracy, and ultimate goals are different. PDA and SDA processes are used for mid- to long-term analyses that help decision-makers estimate required financial needs and assistance to achieve complete recovery. However, an RDA process is used for deciding interventions and priorities in the immediate aftermath of a disaster. An RDA provides local government with the necessary information for an adequate response to life-threatening situations; directs first responders; delivers a quick analysis of the potential hazard to critical infrastructure; determines the need for additional resources; and assists with determining local resource allocations and the need for state and/or federal disaster declaration requests. 


Despite its criticality, conducting a successful PDDA, and particularly RDA, remains difficult. Conventional damage assessment approaches often rely on extensive manual field surveys~\cite{parisi2013earthquake,goretti2002overview}, which can be impractical due to time constraints and the urgency of critical decision-making during post-disaster emergency operations~\cite{macchiarulo2024integrating}. Recent advancements in intelligent data collection systems through emerging technologies, including unmanned aerial vehicles (UAVs)~\cite{khankeshizadeh2024novel}, equipped with advanced sensing mechanisms such as LiDAR~\cite{axel2017building}, thermal imaging~\cite{messina2020applications}, multispectral and hyperspectral cameras~\cite{di2022spectral}, photogrammetry~\cite{jimenez2021digital}, and ground-penetrating radar~\cite{lopez2022unmanned}, have significantly enhanced the ability to rapidly assess post-disaster conditions. An increasing number of studies in recent years have focused on using these technologies for data collection~\cite{al2024integrating,joshi2017damage,yamazaki2007remote,wang2015knowledge} and developing advanced deep learning methods to interpret the collected data~\cite{rastiveis2015building,adriano2021learning,cheng2021deep}. The focus of these studies is on "how" to collect and analyze the data in the aftermath of a disaster. That is, these methods mainly focus on identifying the level of damage to a given infrastructure system using sensing and machine learning technologies. 
Despite the contributions of these studies to the body of knowledge on intelligent PDDA, a critical operational challenge persists: determining "where" to collect the data (where to be observed) to improve the overall assessment of the impacted region, particularly in rapid damage assessments where time is limited. Existing trajectory planning methods for PDDA agents rely solely on prior static data and lack the flexibility and agility needed to adjust data collection targets~\cite{debnath2024review}. 
Therefore, designing and developing adaptive data sampling strategies is essential to fully leverage intelligent, technology-enhanced data collection mechanisms for effective, rapid damage assessments. Such an adaptive approach identifies the most informative locations to be visited and provides a more accurate understanding of the impacted region within a limited time frame.

Formulating an adaptive sampling strategy that overcomes this bottleneck introduces three main challenges. First, post-disaster damage is often spatially heterogeneous and influenced by diverse building features, requiring models that can capture both complex spatial relationships and feature-dependent damage patterns. Second, damage levels are typically expressed on ordinal scales, making it necessary to model ordered categorical outcomes and to learn critical thresholds for identifying regions of interest. Third, adaptive sampling must operate under severe data sparsity and uncertainty while respecting practical operational constraints, such as travel distance, time, and resource expenditure, in order to support cost-effective and informative data collection.

\begin{figure}[t]
  \centering
  \includegraphics[width=\linewidth]{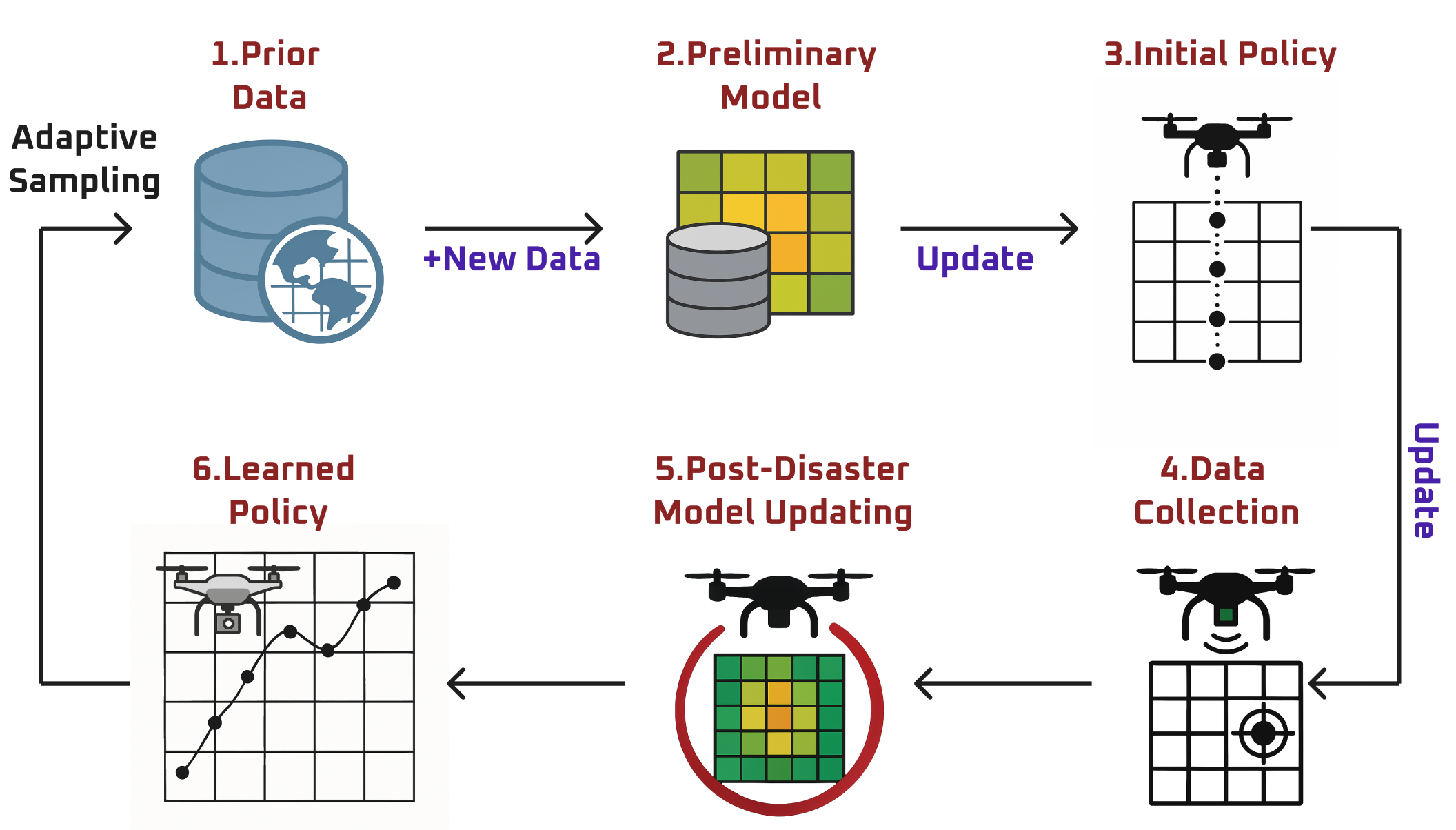}
  \caption{The Proposed Framework for Adaptive Sampling.}
  \label{fig:framework}
\end{figure}

We propose an adaptive sampling framework based on a cost-aware Bayesian optimization strategy. Our core meta-model, the Ordinal Deep Kernel Gaussian Process (ODGP), employs a deep-kernel architecture to capture complex spatial/feature correlations and an ordinal regression formulation to handle categorical damage scales. As illustrated in Figure \ref{fig:framework}, the framework iteratively guides data collectors (e.g., UAVs) using a cost-aware level-set acquisition function. This function identifies locations offering maximum information relative to travel costs, dynamically updating damage predictions and uncertainty estimates to prioritize high-impact regions.


The main contributions of this paper are summarized as follows: (i) Novel Surrogate Model for Ordinal Damage: We introduce an Ordinal Deep Kernel Gaussian Process (ODGP) that effectively captures complex spatial correlations and architectural features while naturally handling discrete, naturally ordered damage levels (e.g., minor, moderate, severe, destroyed); (ii) Cost-Aware Level-Set Acquisition Function: We design a dynamic sampling strategy that simultaneously balances three critical factors: exploration (visiting highly uncertain areas), exploitation (identifying critical damage threshold boundaries via level-set estimation), and travel cost (penalizing long flight trajectories). This enables the UAV to prioritize high-value informational areas without depleting limited battery resources; and (iii) Comprehensive Empirical Validation: We validate the proposed framework through both a controlled synthetic environment and a high-fidelity simulated tsunami scenario. The results demonstrate that our approach achieves highly accurate damage boundary identification while significantly reducing the cumulative flight cost compared to baseline sampling strategies.

\section{Related Work}
\label{sec:literature}

This section reviews two key strands of literature that inform our proposed approach: advances in post-disaster damage assessment (PDDA) and adaptive sampling techniques. Together, these domains motivate our cost-aware Bayesian optimization framework designed to address spatial data collection efficiency under practical operational constraints.

Rapid PDDA is essential for emergency response, resource allocation, and recovery planning. Recently, PDDA has advanced significantly through the integration of remote sensing and machine learning. Various sensors—including satellite imagery~\cite{joshi2017damage,wang2015knowledge}, LiDAR~\cite{rastiveis2015building}, and multi-sensor fusions~\cite{adriano2021learning,al2024integrating,yamazaki2007remote}—provide critical structural and environmental data across affected regions. Concurrently, deep learning architectures, ranging from Convolutional Neural Networks (CNNs)~\cite{cheng2021deep,braik2025post} to hierarchical transformers~\cite{kaur2023large}, have vastly improved damage classification and mapping accuracy. Researchers have also integrated these vision-based assessments with restoration models and agent-based simulations to evaluate broader community resilience~\cite{braik2025framework}. However, these studies primarily focus on "how" to assess damage from existing static datasets, rather than "where" to strategically deploy agents to collect data in a dynamic, resource-limited environment.

To address the challenge of data acquisition, recent efforts have explored adaptive sampling strategies to guide sensing agents toward high-uncertainty or severely damaged areas. While active learning has been used for tasks like flood mapping~\cite{lee2024improving}, it often struggles with distinguishing complex spectral patterns. Reinforcement learning (RL) has been applied to optimize UAV trajectories for post-disaster network recovery~\cite{zhao2022uav}; however, its reliance on extensive simulated pre-training limits its transferability to highly uncertain, real-world disaster landscapes. Bayesian optimization (BO) is highly effective for resource-intensive data collection because it inherently quantifies uncertainty~\cite{diessner2024development}. Yet, in the current PDDA literature, BO is predominantly utilized for hyperparameter tuning or feature selection~\cite{liang2024enhancing,liang2019image}, rather than for spatial trajectory planning and identifying sampling locations.

Despite notable progress in automated PDDA, spatial adaptive sampling remains critically under-explored. The absence of location-targeting strategies limits the operational effectiveness of PDDA agents. To bridge this gap, our study introduces a cost-aware BO framework tailored to enable accurate, adaptive, and resource-efficient data collection in real-world disaster settings.

\section{Methodology.} \label{sec:method}

\begin{figure}[]
  \centering
  \includegraphics[width=\linewidth]{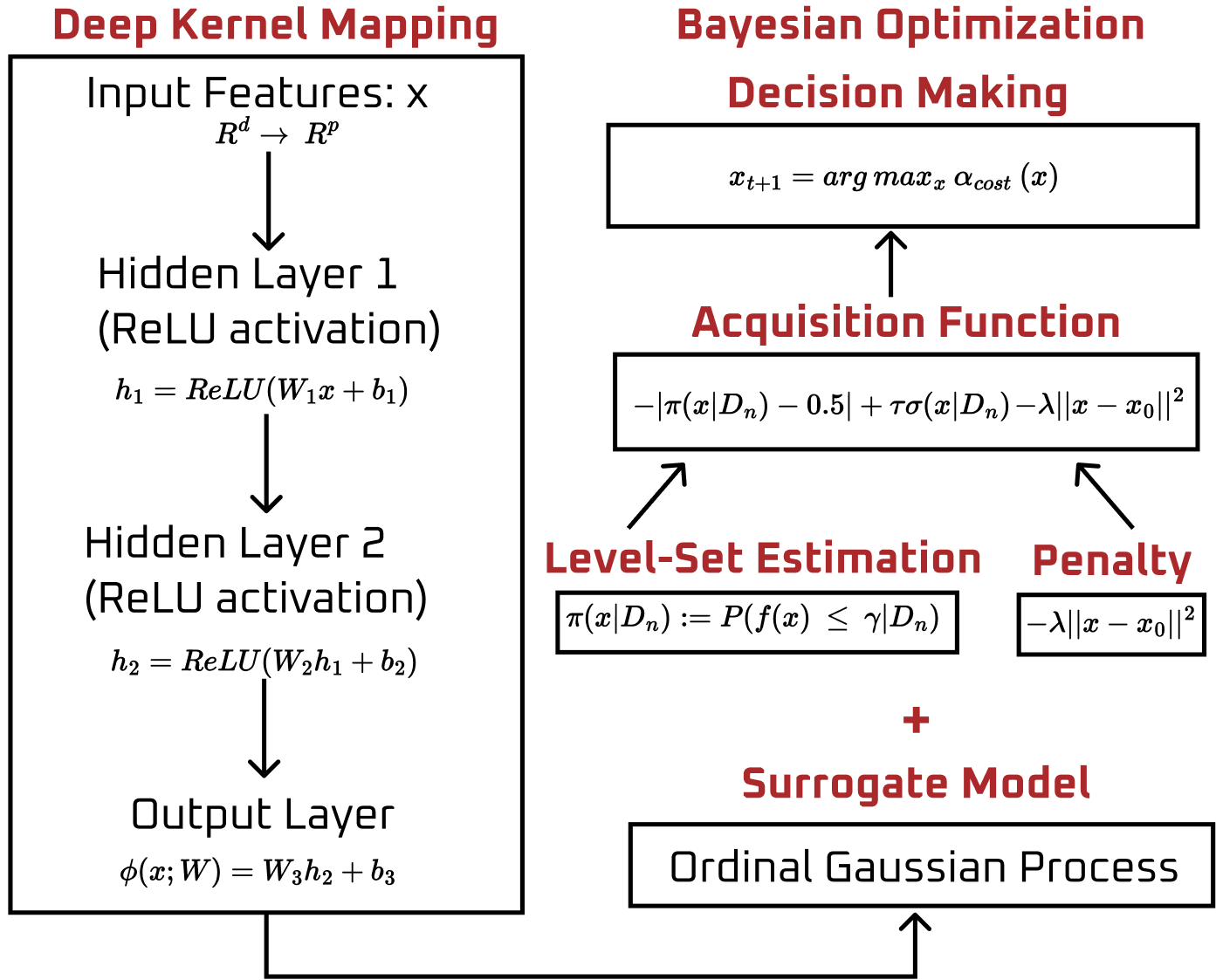}
  \caption{Proposed BO Framework.}
  \label{fig:method}
\end{figure}

We propose a Bayesian optimization (BO) framework to efficiently identify optimal observation locations ("where" to observe) in post-disaster environments. Balancing exploration and exploitation~\cite{shahriari2015taking}, our framework (Figure~\ref{fig:method}) employs an Ordinal Deep Kernel Gaussian Process (ODGP) as its surrogate model. The ODGP's deep-kernel architecture captures \textit{complex spatial and feature-dependent damage patterns} (Challenge 1), while its ordinal formulation natively accommodates severity ratings to enable the \textit{identification of critical regions of interest (ROI)} (Challenge 2). To mitigate data sparsity under \textit{operational constraints} (Challenge 3), we design a cost-aware acquisition function coupled with level-set estimation. By penalizing excessive travel distances, this guides BO toward sampling locations that are both maximally informative and cost-efficient.

The remainder of this section is structured as follows: We first introduce standard GPs and the deep-kernel learning used for feature embedding. Next, we formulate the proposed ODGP, followed by the design of the cost-aware acquisition function and level-set estimation. Finally, we detail the variational inference procedure for joint parameter optimization.

\subsection{Gaussian Process.}
Gaussian processes are widely used as surrogate models in Bayesian optimization to approximate unknown objective functions. Formally, a GP defines a distribution over functions such that any finite subset of observations (or function evaluations) follows a joint Gaussian distribution \cite{10360364}.
\paragraph{GP's Prior.} Let $X_n=[\mathbf{x}_1^\top,\ldots,\mathbf{x}_n^\top]^\top\in\mathbb{R}^{n\times d}$ denote the training inputs, and let $\mathbf{y}_n=[y_1,\ldots,y_n]^\top$ be the corresponding observations. We assume that the outputs are noisy evaluations of an unknown latent function $f:\mathbb{R}^{d} \rightarrow \mathbb{R}$, such that $\mathbf{y}_n=f(X_n)+\boldsymbol{\epsilon}_n$,
where $\boldsymbol{\epsilon}_n$ refers to the noise and is assumed to follow a normal distribution with $\boldsymbol{\epsilon}_n \sim \mathcal{N}(0,\sigma^2 I_n)$.
We place a Gaussian process prior on the function $f$, denoted as:
\begin{equation}
    f \sim \mathcal{GP}(m(\cdot),k(\cdot,\cdot)),
\end{equation}
where $m(\cdot)$ is the mean function and $k(\cdot,\cdot)$ is a kernel (covariance) function. 
This prior expresses the belief that, for any finite collection of inputs, the corresponding function values follow a joint Gaussian distribution. In particular, under the GP prior, the vector of latent function values: $\mathbf{f}_n = f(X_n)= \begin{bmatrix} f(\mathbf{x}_1), f(\mathbf{x}_2), \cdots, f(\mathbf{x}_n) \end{bmatrix}^\top$ for a set of inputs $X_n$, is distributed as a multivariate normal distribution: $\mathbf{f}_n \sim \mathcal{N}(\mathbf{m}_X,\mathbf{K}_{XX})$, where $\mathbf{m}_X=m(X_n)$ is the vector of prior means and $[\mathbf{K}_{XX}]_{i,j}=k(\mathbf{x}_i,\mathbf{x}_j)$ is the covariance matrix defined by the kernel.
\paragraph{GP's Posterior.} Given the noisy observations $\mathbf{y}_n$, we can derive the posterior distribution over the latent function values $\mathbf{f}_n$ using Bayes' rule. The posterior is also Gaussian: $\mathbf{f}_n|X_n,\mathbf{y}_n \sim \mathcal{N}(\boldsymbol{\mu}_f,\Sigma_f)$, with:
\begin{align}
\boldsymbol{\mu}_f &=\mathbf{m}_X+\mathbf{K}_{XX}(\mathbf{K}_{XX}+\sigma^2I_n)^{-1}(\mathbf{y}_n-\mathbf{m}_X),\\
\Sigma_f &= \mathbf{K}_{XX} - \mathbf{K}_{XX}(\mathbf{K}_{XX}+\sigma^2I_n)^{-1} \mathbf{K}_{XX}.
\end{align}
This posterior quantifies the updated belief about the latent function after observing data. It captures both the predictive mean and the uncertainty associated with each prediction.
\paragraph{Prediction.} To make predictions on new inputs $X_* = [\mathbf{x}_1^*, \ldots, \mathbf{x}_{n_*}^*]^\top \in \mathbb{R}^{n_* \times d}$, we consider the posterior predictive distribution of the corresponding latent values $\mathbf{f}_*=f(X_*)$, conditioned on the training data $(X_n,\mathbf{y}_n)$. This predictive distribution is also Gaussian: $ \mathbf{f}_*|X_n,\mathbf{y}_n,X_* \sim \mathcal{N}(\boldsymbol{\mu}_*, \Sigma_*)$, where the predictive mean and covariance are given by:
\begin{align}
    \boldsymbol{\mu}_* &= \mathbf{m}_* + \mathbf{K}_{*X}(\mathbf{K}_{XX} + \sigma^2 I_n)^{-1}(\mathbf{y}_n - \mathbf{m}_X), \\
    \Sigma_* &= \mathbf{K}_{**} - \mathbf{K}_{*X}(\mathbf{K}_{XX} + \sigma^2 I_n)^{-1} \mathbf{K}_{X*}.
\end{align}
Here, \( \mathbf{m}_* = m(X_*) \) is the vector of prior means at the test locations, \( \mathbf{K}_{*X} \) is the \( n_* \times n \) cross-covariance matrix between the test and training inputs, \( \mathbf{K}_{**} \) is the \( n_* \times n_* \) covariance matrix at the test inputs, and \( \mathbf{K}_{X*} = \mathbf{K}_{*X}^\top \).

\subsection{Proposed Ordinal Deep Kernel Gaussian Process.}
\label{sec:ordinal regression}
In conventional GP regression, two key gaps limit performance on post-disaster damage assessment. First, the kernel functions that use the distance between the raw geographic or building attributes $\mathbf{x} \in \mathbb{R}^d$ may fail to capture complex patterns that underlie damage severity, limiting the model's ability to make accurate predictions. Second, the GP regression typically assumes that the observed training labels represent continuous variables. However, in many practical scenarios, particularly in damage assessment tasks, these labels are \emph{ordinal} categories (e.g., "no damage", "minor", "moderate", "severe", "destruction") that imply a ranked structure without precisely quantifiable numerical distances. To address these two challenges, we propose an \emph{Ordinal Deep Kernel Gaussian Process} model that extends the standard GP framework in two ways: it optionally replaces the raw input features with a learned neural embedding to capture complex feature patterns \cite{wilson2016stochastic}, and it models the output through a latent variable linked to discrete ordinal categories by learning a set of thresholds. We explain these two components in the following sections.

\paragraph{Deep Kernel Learning.}
\label{sec:dkl}
To capture rich patterns in the input-output space, we introduce a neural embedding that transforms feature vector \(\mathbf{x}\) into a latent representation: $\phi(\mathbf{x}; \mathbf{W}): \mathbb{R}^d \;\longrightarrow\; \mathbb{R}^p$. Specifically, we parametrize the mapping simply by a two‐layer multilayer perceptron:
\begin{equation}\label{eq:neural-feature-mapping}
\begin{aligned}
    &\mathbf{h}_1 = \mathrm{ReLU}\!\bigl(\mathbf{W}_1\,\mathbf{x} + \mathbf{b}_1\bigr), \\
    &\mathbf{h}_2 = \mathrm{ReLU}\!\bigl(\mathbf{W}_2\,\mathbf{h}_1 + \mathbf{b}_2\bigr), \\
    &\phi(\mathbf{x}; \mathbf{W}) = \mathbf{W}_3\,\mathbf{h}_2 + \mathbf{b}_3,
\end{aligned}
\end{equation}
where $\mathrm{ReLU(.)}$ denotes the Rectified Linear Unit function defined as $\mathrm{ReLU}(x)=max\{0,x\}$, and $\mathbf{W}=\{\mathbf{W}_1,\mathbf{b}_1,\mathbf{W}_2,\mathbf{b}_2,\mathbf{W}_3,\mathbf{b}_3\}$ collects all trainable weights and biases. Optional batch normalization and dropout layers may be inserted after each hidden layer to stabilize training and improve generalization. The architecture of this neural network can be modified depending on the complexity of the problem. 

We assume a GP prior over the latent function $f$ defined on the neural embedding $\phi(\mathbf{x}; \mathbf{W})$. This deep-kernel construction enables joint learning of the neural embedding and the GP within a unified model:

\begin{equation}
f
\sim
\mathcal{GP}\!\left(
m(\cdot),
k\!\left(
\phi(\mathbf{x};\mathbf{W}),
\phi(\mathbf{x}';\mathbf{W})
\right)
\right).
\end{equation}

\paragraph{Ordinal Log-Likelihood.} To link the latent GP output to the discrete ordinal result, we introduce an ordinal log-likelihood. We define the output of the deep-kernel GP, $g(\mathbf{x})=f\left(\phi\left(\mathbf{x}_i ; \mathbf{W}\right)\right)$, as a continuous risk score representing the underlying damage severity for input $\mathbf{x}_i$. This continuous score is then mapped to an ordinal label $y_i^{\mathcal{O}} \in\{1, \ldots, \mathcal{C}\}$ using a set of learnable thresholds denoted by $\mathbf{t}=\left\{t_1, t_2, \ldots, t_{\mathcal{C}-1}\right\}$ so that:
\begin{equation}
y_i^{\mathcal O}
=
1 + \sum_{c=1}^{\mathcal C-1}
\mathbf{1}\!\left(g(\mathbf{x}_i)>t_c\right).
\end{equation}
where $\mathbf{1}$ is the indicator function, and the thresholds satisfy
$t_0=-\infty$,
$\qquad
t_{\mathcal C}=+\infty$,
$\qquad
t_1<\cdots<t_{\mathcal C-1}$.
The probability of the label belonging to ordinal category $c$, conditioned on input $\mathbf{x}_i$, is:

\begin{equation}
\begin{aligned}
\mathbb{P}(y_i^{\mathcal O}=c\mid\mathbf x_i)
&=\Phi\bigl(t_c - f(\phi(\mathbf x_i;\mathbf W))\bigr)\\
&\quad-\;\Phi\bigl(t_{c-1} - f(\phi(\mathbf x_i;\mathbf W))\bigr)\,,
\end{aligned}
\end{equation}
where $\Phi(\cdot)$ is the standard normal cumulative distribution function (CDF). 
Consequently, the contribution of sample $i$ to the log-likelihood becomes:

\small
\begin{equation}
\begin{aligned}
\log \mathcal{L}\left(\mathbf{W},\mathbf{t}|\mathbf{x}_i,y_i^\mathcal{O}\right)=\sum_{c=1}^{\mathcal{C}}\left[y_i^\mathcal{O}=c\right]\log\left(\mathbb{P}(y_i^{\mathcal O}=c\mid\mathbf x_i)\right),
\end{aligned}
\end{equation}
\normalsize
where $\mathbf{W}$ and $\mathbf{t}$ are deep-kernel learning parameters and learnable thresholds, respectively. The Iverson bracket notation $[P]$ denotes an indicator function, and the expression is set to one if $P$ is true and zero otherwise \cite{iverson1962programming}. This ordinal likelihood structure, combined with the deep-kernel embedding $\phi(\mathbf{x} ; \mathbf{W})$, allows the model to learn the input representations ($\mathbf{W}$), kernel hyperparameters, and category thresholds ($\mathbf{t}$) jointly within the variational training framework described in Section \ref{sec:training}. Consequently, the trained ODGP model provides the two crucial outputs needed for the adaptive sampling stage:
1) A posterior distribution (i.e., mean and variance) over the continuous risk score $g(\mathbf{x})$ for any given location.
2) The learned set of damage thresholds $\mathbf{t}$ that denote the different levels of damage. As detailed in Section \ref{sec:acquisition}, these two components are used directly by the Bayesian optimization acquisition function to guide the next sampling decision.

\subsection{Proposed Cost-Aware Level-Set Acquisition for Bayesian Optimization.}
\label{sec:acquisition}
In Bayesian optimization, the acquisition function is pivotal to the sequential sampling process, as it quantifies the expected utility of evaluating the black-box function at each candidate input and thereby guides the selection of new sample locations. It balances the exploitation of regions with high predicted performance and the exploration of regions with high epistemic uncertainty. Classical acquisition functions, such as Expected Improvement (EI) \cite{movckus1975bayesian} and Upper Confidence Bound (UCB) \cite{srinivas2009gaussian}, are typically designed to locate global optima. However, in our setting, the goal is not to find maxima or minima of the underlying function, but rather to accurately characterize regions where the response exceeds a predefined critical threshold. Moreover, in many real-world applications, such as UAV-based post-disaster assessment, evaluations are subject to spatial mobility constraints and resource limitations. Arbitrarily sampling across the domain may incur substantial operational costs due to energy consumption, travel time, or mission feasibility.

To address these challenges, we propose a novel \textit{Cost-Aware Level-Set Acquisition Function} powered by our ODGP model. To make the formulation concise, we define the continuous risk score as $g(\mathbf{x}):=$ $f(\phi(\mathbf{x} ; \mathbf{W}))$. Our approach is similar to the straddle criterion for level-set estimation (LSE), which refers to the task of identifying regions of the input space where an unknown function lies above or below a specific threshold \cite{bryan2005active}. 
Given the dataset $\mathcal{D}_n=\left\{\left(x_i, y_i^{\mathcal{O}}\right)\right\}_{i=1}^n$ and the agent's current location $\mathbf{x}_{\text {curr }}$, the next sampling point is chosen by maximizing the following cost-aware acquisition function:
\begin{equation}\label{eq:straddle}
\begin{aligned}
\alpha_{\mathrm{cost}}(\mathbf{x}) =& -\left|\pi_n(\mathbf{x}\mid\mathcal{D}_n) - 0.5\right| \\
& + \tau\,\sigma_n(\mathbf{x}\mid\mathcal{D}_n) - \lambda\,\bigl\|\mathbf{s}(\mathbf{x}) - \mathbf{s}(\mathbf{x}_{\mathrm{curr}})\bigr\|^2,
\end{aligned}
\end{equation}
where $\pi_n(\mathbf{x}\mid\mathcal{D}_n)$ is the level-set posterior probability and is defined and computed as 
\begin{equation}
\begin{aligned}
\pi_n(\mathbf{x}\mid\mathcal D_n)
&=
\mathbb P\!\left(
g(\mathbf{x})\geq\gamma
\mid\mathcal D_n
\right)
=
1-\Phi\!\left(
\frac{\gamma-\mu_n(\mathbf{x})}
{\sigma_n(\mathbf{x})}
\right).
\end{aligned}
\end{equation}
This probability quantifies the model's confidence that a candidate input lies within the critical damage region $\mathcal{L}_{\gamma}^{+}(g)=\{\mathbf{x}:g(\mathbf{x})\geq\gamma\}$ where $\gamma$ refers to a specific damage level threshold. A key advantage of our framework is that this threshold $\gamma$ is not chosen arbitrarily. Instead, it is directly obtained from the set of thresholds $\mathbf{t}=\left\{t_1, \ldots, t_{\mathcal{C}-1}\right\}$ learned by the ODGP model. 
This provides a principled, data-driven way to define the ROI. In addition, $\sigma_n(\mathbf{x}\mid\mathcal{D}_n)$ is the posterior standard deviation. This formulation is conceptually similar to the widely-used UCB framework, as it additively combines an exploitation term with an exploration term ($\tau \sigma_n(\mathbf{x})$). The key distinction lies in our exploitation term, which is specifically designed for the LSE goal of boundary refinement rather than global optimization. This composite criterion intelligently balances three key objectives: (1) \textit{Exploitation}: The first term,$-\left|\pi_n(\mathbf{x})-0.5\right|$ is maximized when the posterior probability of a point belonging to the target region, $\pi_n(\mathbf{x})$, is exactly $0.5$, which is the point of maximum classification uncertainty; (2) \textit{Exploration}: The second term, $\tau \sigma_n(\mathbf{x})$, encourages exploration of regions with high epistemic uncertainty (given by the posterior standard deviation $\sigma_n(\mathbf{x})$ ). This promotes global learning, and the hyperparameter $\tau>0$ modulates the exploration-exploitation trade-off; (3) \textit{Cost Penalty}: The final term, $-\lambda\left\|\mathbf{s}(\mathbf{x})-\mathbf{s}(\mathbf{x}_{\mathrm{curr}})\right\|^2$, is an additive penalty that discourages costly spatial transitions. Here $\mathbf{s}(\mathbf{x})$ extracts only the geographic coordinates of a candidate (latitude and longitude), so that non-spatial attributes such as year built or floor area do not enter the travel penalty. In our experiments these coordinates are standardized to zero mean and unit variance over the candidate set before the squared Euclidean distance is computed, yielding a dimensionless proxy for relative travel cost (e.g., energy consumption, travel time); the hyperparameter $\lambda>0$ controls the strength of this penalty.  

Following the same rule in BO, maximizing \(\alpha_{\mathrm{cost}}(\mathbf{x})\) gives the next sampling point:
\begin{equation}
    \mathbf{x}_{n+1} = \arg\max_{\mathbf{x}} \alpha_{\mathrm{cost}}(\mathbf{x}),
\end{equation}
which jointly promotes boundary refinement of the target superlevel set \(\mathcal{L}_{\gamma}^{+}(g)\) and efficient navigation of the domain—ensuring both informational gain and operational feasibility.

\subsection{Model Training.}
\label{sec:training}
Exact inference in GP models becomes intractable when paired with non-Gaussian likelihoods, such as ordinal likelihood used in our setting, since it lacks a closed-form posterior. Moreover, even for models with Gaussian likelihoods, the computational cost of exact inference scales cubically with the number of training points $n$, making it prohibitive for large datasets. To overcome both the intractability of inference in ordinal GP models and the scalability limitations of standard GP inference, we adopt the sparse variational Gaussian process (SVGP) framework \cite{hensman2015scalable}. SVGP provides a principled and efficient approximation by introducing a set of inducing variables that summarize the function, enabling variational inference and scalable training via stochastic optimization. Below, we detail the main components of this framework and the end-to-end optimization procedure.

\paragraph{Inducing Points and Variational Posterior.} 
The central idea behind SVGP is to approximate the full GP posterior using a set of \emph{inducing points}. Let $\mathbf{Z} = [\mathbf{z}_1^\top, \ldots, \mathbf{z}_m^\top]^\top \in \mathbb{R}^{m \times d}$ denote the set of $m<n$ inducing points, and let $\mathbf{u} = f(\mathbf{Z}) \in \mathbb{R}^m$ be the corresponding latent values (e.g., latent damage scores). We place a variational distribution over $\mathbf{u}$ as follows:
\begin{equation}
    q(\mathbf{u})
=
\mathcal{N}(\mathbf{u}\mid\mathbf{m},\mathbf{S}).
\end{equation}
where $\mathbf m \in \mathbb{R}^m$ is the variational mean and $\mathbf S \in \mathbb{R}^{m \times m}$ is the variational covariance matrix. In practice, to guarantee numerical stability and positive-definiteness, we parameterize $\mathbf S$ by its Cholesky factor matrix $\mathbf{L}$, such that $\mathbf S = \mathbf L\mathbf L^\top$, where $\mathbf L$ is lower-triangular.

\paragraph{Conditional Distribution at Training Inputs.}
Given training inputs $X_n = [\mathbf{x}_1^\top, \ldots, \mathbf{x}_n^\top]^\top$ and their latent function values $\mathbf{f}=[f(\mathbf{x}_1),...,f(\mathbf{x}_n)]^T$, the GP prior yields the conditional distribution:
\begin{equation} \label{conditional}
    p(\mathbf f \mid \mathbf u) = \mathcal{N}(\mathbf f \mid \mathbf{A}_{nm}\mathbf{u},\; \mathbf{K}_{xx} - \mathbf{A}_{nm}\mathbf{K}_{uu}\mathbf{A}_{nm}^\top),
\end{equation}
where $\mathbf{K}_{xx} \in \mathbb{R}^{n \times n}$ is the kernel matrix evaluated at the training input; $\mathbf{K}_{uu} \in \mathbb{R}^{m \times m}$ is the kernel matrix evaluated at the inducing points $Z$; $\mathbf{K}_{xu} \in \mathbb{R}^{n \times m}$ is the cross-covariance matrix between $X$ and $Z$; and $\mathbf{A}_{nm} = \mathbf{K}_{xu} \mathbf{K}_{uu}^{-1}$ expresses the linear relationship between $\mathbf f$ and $\mathbf u$. 

Intuitively, Eq. \eqref{conditional} quantifies how well the inducing variables $\mathbf{u}$ explain the latent GP values $\mathbf{f}$ at the actual training inputs $X$. Specifically, the mean term $\mathbf{A}_{nm}\mathbf{u}$ predicts the GP values at the training points by interpolating from the inducing values $\mathbf{u}$. The correction term $\mathbf{K}_{xx} - \mathbf{A}_{nm}\mathbf{K}_{uu}\mathbf{A}_{nm}^\top$ adds a correction that quantifies the residual uncertainty when the training points lie away from the inducing set.

\paragraph{Marginal Variational Distribution over Training Function Values.}
To complete the variational inference procedure and to make predictions or evaluate the likelihood of the data, we need the marginal distribution of $\mathbf{f}$ by integrating out the inducing variables:
\begin{equation}\label{marginal}
    q(\mathbf f) = \int p(\mathbf f \mid \mathbf u)\,q(\mathbf u)\,d\mathbf u = \mathcal{N}(\mathbf f \mid \boldsymbol\mu_f, \boldsymbol\Sigma_f),
\end{equation}
with $\boldsymbol\mu_f = \mathbf{A}_{nm}\mathbf{m}$; $\boldsymbol\Sigma_f = \mathbf{K}_{xx} - \mathbf{A}_{nm}\mathbf{K}_{uu}\mathbf{A}_{nm}^\top + \mathbf{A}_{nm}\mathbf{S}\mathbf{A}_{nm}^\top.$ Here, the mean term $\boldsymbol\mu_f$ can be interpreted as the best estimate of the latent function at each training input given the variational mean at the inducing points. The covariance $\boldsymbol\Sigma_f$ reflects the overall uncertainty, combining the original GP prior uncertainty with the additional uncertainty due to the variational posterior.

\paragraph{Variational Objective and Optimization.}
The entire model's parameters are trained by maximizing the evidence lower bound (ELBO) \cite{kingma2019introduction}. The ELBO is derived by marginalizing out $\mathbf{u}$ via $q(\mathbf{f})$ in Eq. \eqref{marginal}, and it balances two objectives: (i) The data likelihood term $\mathbb{E}_{q(f_i)}\bigl[\log p(y_i^O\mid f_i)\bigr]$, computed through $q(\mathbf{f})$ (which depends on $q(\mathbf{u})$), ensures fit to the observed data; (ii) The Kullback-Leibler (KL) divergence term $\mathrm{KL}\bigl[q(\mathbf u)\|\;p(\mathbf u)\bigr]$ regularizes the variational distribution over inducing points, enforcing smoothness inherited from the GP prior~\cite{kullback1951information}.
The trade-off between these two objectives, automatically balanced during optimization, ensures that the model flexibly adapts to data while avoiding overfitting. The ELBO objective is defined as:

\begin{equation}
\begin{aligned}
\mathcal{L}_{\mathrm{ELBO}}
= \sum_{i=1}^n \mathbb{E}_{q(f_i)}\bigl[\log p(y_i^O\mid f_i)\bigr]- \mathrm{KL}\bigl[q(\mathbf u)\|\;p(\mathbf u)\bigr]. 
\end{aligned}
\end{equation}

In summary, the proposed cost-aware BO framework operates iteratively. At each step, the ODGP model is updated using the available observations by maximizing the ELBO (Section \ref{sec:training}). Then, the cost-aware acquisition function (Eq. \ref{eq:straddle}) is evaluated across the candidate space to dictate the next sampling location. Once the new observation is acquired, it is appended to the dataset, and the process repeats until the operational budget is exhausted. This closed-loop process is visually summarized in Figure \ref{fig:method}.

\section{Experiments and Results.}
\label{sec:experiments}
In this section, we evaluate the performance of our proposed cost-aware data collection framework. To systematically demonstrate its effectiveness, the evaluation is divided into two parts. First, we use a controlled 2D synthetic example to intuitively illustrate how the cost-aware acquisition function guides the sampling trajectory and reduces uncertainty. Second, we apply the framework to a high-fidelity simulated tsunami damage assessment in Seaside, Oregon, demonstrating the scalability and robustness of the deep-kernel learning approach.

\subsection{Performance Evaluation using Toy Example.}
\label{sec:toy}
We first used the toy study to demonstrate how the proposed method can achieve accurate damage prediction and the evolution of the agent's uncertainty during the adaptive sampling process.

\begin{figure*}[!htbp]
  \centering
  \includegraphics[width=0.8\textwidth]{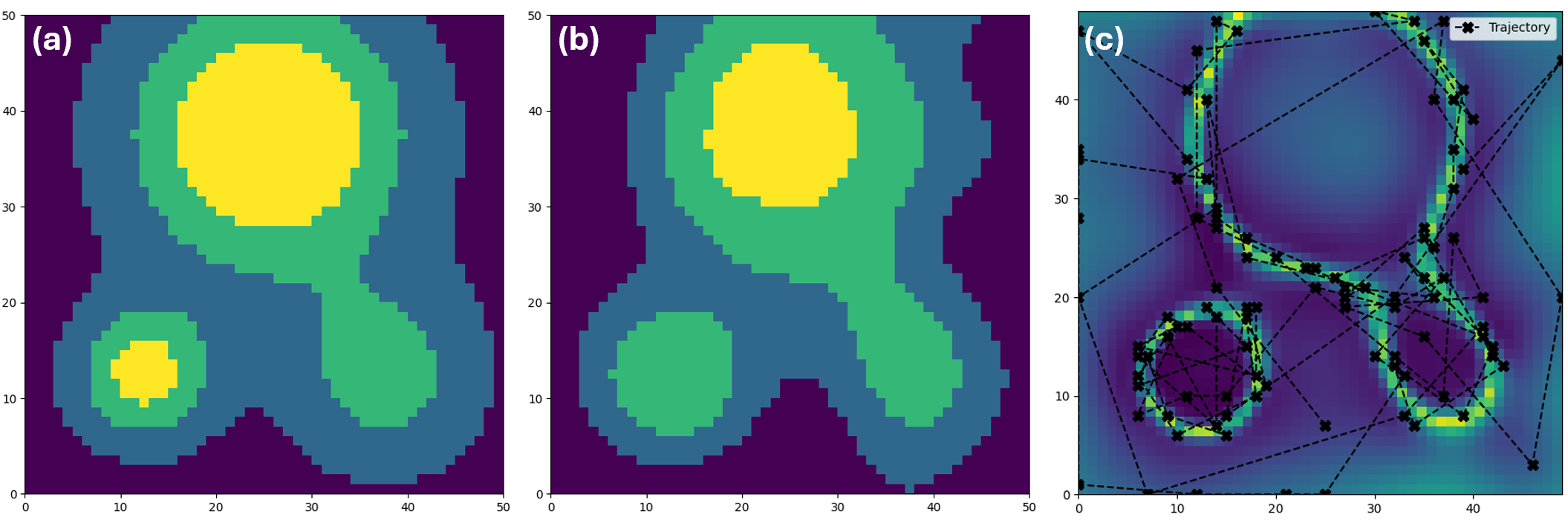}
  \caption{Experiment Results From the Synthetic Example. (a) Ground Truth Damage Map; (b) Model Predicted Map; (c) Sampling Trajectory.}
  \label{fig:toy_result}
\end{figure*}

\paragraph{Data Generation.}
We generate a synthetic "damage" map by superposing three smooth Gaussian hotspots on an $L\times L$ grid. Each Gaussian hotspot is defined by a center $\mu_k$, spread $\sigma_k$, and peak intensity $p_k$. The continuous field is then clamped to $[0,C]$ and rounded to the nearest integer to yield four ordinal damage levels. Concretely, for each grid cell $(i,j)\in\{0,\dots,L-1\}^2$, we define
\begin{equation}
\label{eq:toy_damage}    
\tilde f(i,j)
= \left\lfloor [F(i,j)]_{[0,C]} + \tfrac12 \right\rfloor,
\qquad
\end{equation}
with
\begin{equation}
F(i,j)
= \sum_{k=1}^{K} p_k\exp\left(-\frac{\|(i,j)-\mu_k\|^2}{2\sigma_k^2}\right),
\qquad K=3.
\end{equation}
where $
[x]_{[0,C]} = \min\{\max\{x,0\},C\}$, 
$\mu_k=(x_{0,k},y_{0,k})$, and $\lfloor\,\cdot\,+\tfrac12\rfloor$ rounds to the nearest integer in $\{0,1,...,C\}$. In our setting, $L=50$, $C=3$, and $K=3$ Gaussian anomalies with centers $\mu_k=((12,12);(37,12);(25,37))$; spreads $\sigma_k=(5;7;10)$, and peaks $p_k=(3;2;4)$, are summed, and rounded to $\{0,1,2,3\}$. The generated damage map is shown in Figure \ref{fig:toy_result} (a).  

\paragraph{Model Performance.}
We leverage the proposed method to sequentially collect data from the underlying damage map with the goal of recovering the damage map with the lowest data collection cost. Figure \ref{fig:toy_result} demonstrates (a) the ground truth damage map, (b) the ordinal predictions obtained from our proposed method, and (c) exploration trajectory of the agent, respectively. The resemblance of Figure \ref{fig:toy_result} (a) and Figure \ref{fig:toy_result} (b) demonstrates that the proposed method can recover the underlying damage map (with around $86\%$ accuracy). More specifically, it can identify the borders of each specific damage cluster. In Figure \ref{fig:toy_result} (c), the trajectory of the data collection agent shows that the agent predominantly samples along the boundaries of the damage regions. This behavior is a direct result of the acquisition function, which balances the goal of level-set estimation with predictive uncertainty. By focusing on the boundaries of the region rather than fully penetrating the high-damage zones, the agent efficiently identifies critical areas while minimizing the exploration time, thus improving operational efficiency in time-sensitive scenarios. 

\begin{figure*}[!htbp]
  \centering
  \includegraphics[width=0.9\textwidth]{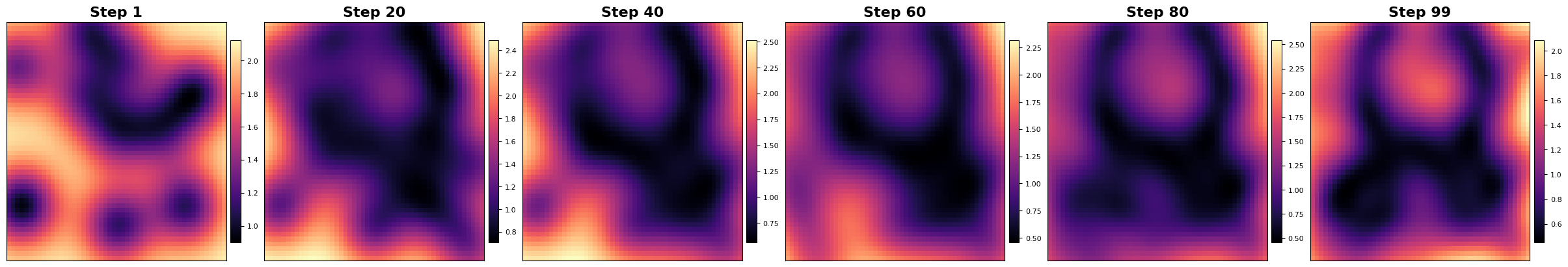}
  \caption{Agent's Uncertainty Map Evolution.}
  \label{fig:gp_variance}
\end{figure*}

\paragraph{Uncertainty Evolution of the Agent.}
In addition, Figure \ref{fig:gp_variance} illustrates how the agent's predictive uncertainty evolves over the exploration steps. Darker regions indicate areas of low predictive uncertainty, suggesting that the agent has visited frequently in these regions, resulting in lower prediction uncertainty. In particular, these low-uncertainty regions align closely with the areas most severely damaged, reflecting the agent's ability to focus on the most critical parts of the impacted region. At step 1, the agent's uncertainty is spatially unstructured (compared to the original damage map) since not many observations are available. As the agent collects new measurements, its posterior belief sharpens: predictive variance decreases most rapidly where data is gathered. By step 99, the map of low variance clearly highlights the critical damage regions, demonstrating the agent's ability to focus its exploration.

\subsection{Tsunami Damage Assessment Case Study.}
\label{sec:result}
In this section, we empirically evaluate the performance of the proposed method using a realistic tsunami scenario generated by the R2D simulator. Performance metrics are accuracy, weighted F$_1$ score, and travel cost.

\paragraph{Dataset Description.}
\label{sec:r2d}
As introduced previously, our empirical evaluation utilizes a simulated 500-year CSZ tsunami scenario in Seaside, Oregon~\cite{frank_mckenna_2025_15446356}. The dataset comprises 1000 buildings, each defined by a feature vector $\mathbf{x}$ with six attributes: latitude, longitude, number of stories, year built, floor area, and GIS-derived area in acres (GIS-ACRES). Only the spatial mapping $\mathbf{s}(\mathbf{x})=(\mathrm{latitude},\mathrm{longitude})$ is used in the travel-cost term of Eq.~\eqref{eq:straddle}; as noted above, these coordinates are standardized over the candidate set. The building-specific damage is classified into four ordinal levels: I (minor), II (moderate), III (severe), and IV (destruction). To simulate a sparse data setting for adaptive assessment, we assume only 10 buildings are initially observed.

\paragraph{Experimental Setup.}
Six strategies (three with ODGP and three with GP) are compared, each using either an ordinal deep-kernel Gaussian process (ODGP) or a standard Gaussian Process (GP) as the surrogate model. Sampling is guided either by an acquisition function (AF) with $\lambda=0$ in Eq.~\eqref{eq:straddle}, a cost-aware acquisition function (CAF) with $\lambda=0.5$, or random sampling (R). The strategies are denoted by \textit{GP-CAF}, \textit{ODGP-CAF}, \textit{GP-AF}, \textit{ODGP-AF}, \textit{GP-R}, and \textit{ODGP-R}. 

\paragraph{Results and Discussion.}

Table~\ref{tab:acc_result} presents the classification performance across 100 sampling episodes. The ODGP variants consistently outperform standard GPs, demonstrating the deep-kernel's superior capability in capturing complex spatial and feature-based correlations. Our proposed \textit{ODGP-CAF} achieves the best overall performance, attaining the highest accuracy ($85\% \pm 1\%$) and weighted F1 score ($0.81 \pm 0.01$). While \textit{ODGP-AF} and \textit{ODGP-R} eventually reach comparable accuracy levels, they do so \textit{at the expense of substantially higher travel costs.}


\begin{table}[tbp]
  \centering
  \caption{Comparison Results on Model Performance at Episode = 100.}
  \label{tab:acc_result}
  \setlength{\tabcolsep}{4pt}   
  \footnotesize
  \begin{tabular}{lccc}
    \toprule
    Algorithm & Accuracy & Weighted F1 & Travel Cost \\
    \midrule
    \textbf{ODGP-CAF} & \textbf{85\% $\pm$ 1\%} & \textbf{0.81 $\pm$ 0.01} & \textbf{44.69 $\pm$ 4.18} \\
    ODGP-AF  & 83\% $\pm$ 3\% & 0.80 $\pm$ 0.02 & 64.14 $\pm$ 7.13 \\
    ODGP-R   & 82\% $\pm$ 2\% & 0.78 $\pm$ 0.02 & 77.92 $\pm$ 7.88 \\
    GP-CAF  & 77\% $\pm$ 2\% & 0.73 $\pm$ 0.02 & 32.89 $\pm$ 3.86 \\
    GP-AF   & 75\% $\pm$ 3\% & 0.71 $\pm$ 0.03 & 46.31 $\pm$ 5.82 \\
    GP-R    & 81\% $\pm$ 2\% & 0.77 $\pm$ 0.02 & 82.46 $\pm$ 4.54 \\
    \bottomrule
  \end{tabular}
\end{table}

\begin{figure*}[!htbp]
  \centering  \includegraphics[width=0.8\linewidth]{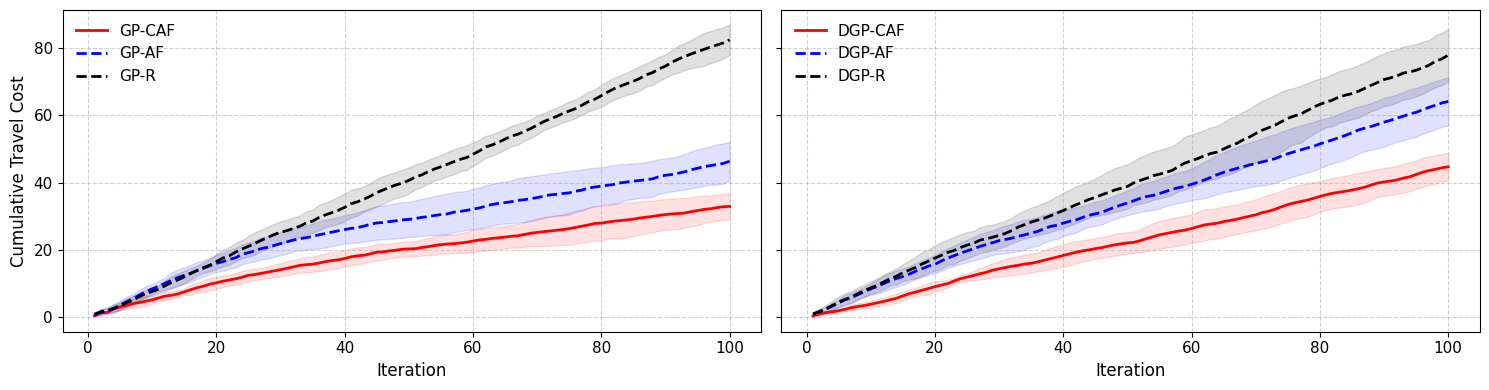}
  \caption{Comparison of Agent Travel Costs. Legend labels ``DGP'' denote the ordinal deep-kernel GP (ODGP) variants.}
  \label{fig:cost}
\end{figure*}

Figure~\ref{fig:cost} further highlights the cumulative travel costs. Across both GP (left) and ODGP (right) families, the cost-aware variants (-CAF) consistently maintain the lowest cumulative costs. This confirms that penalizing travel distance does not compromise predictive accuracy; rather, it significantly improves operational efficiency compared to standard (-AF) or random (-R) guidance.

\begin{figure}[H]
  \centering
  \includegraphics[width=0.9\linewidth]{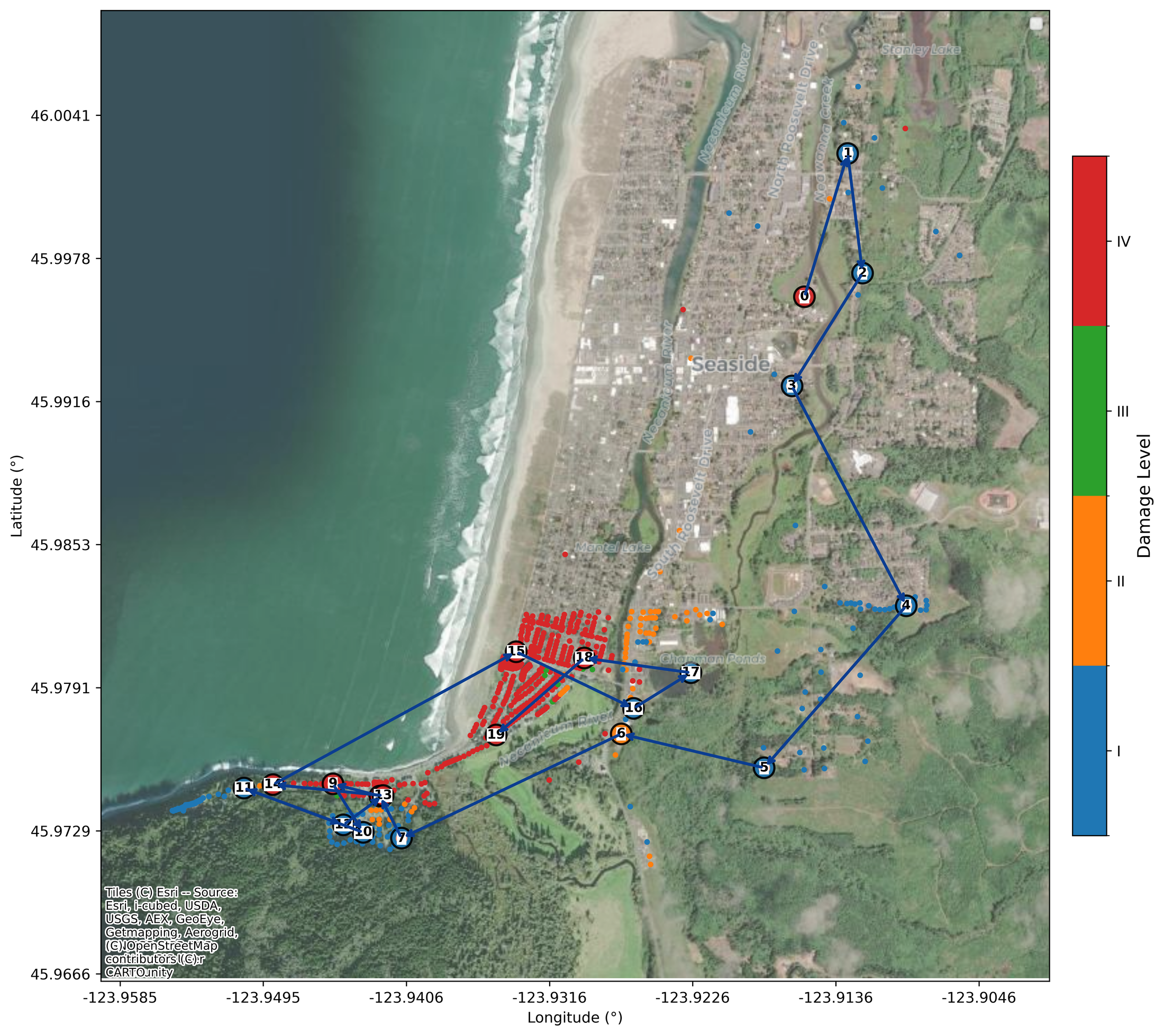}
  \caption{Agent's Learned Action on Tsunami Damage Map. Four ordinal damage levels-(I: blue, II: orange, III: green, IV: red) are color‐coded. Numbered markers denote the agent’s sequential sampling decisions and arrows trace its path.}
  \label{fig:tsuma-traj}
\end{figure}

To understand the agent's behavior, Figure~\ref{fig:tsuma-traj} illustrates the sampling trajectory of \textit{ODGP-CAF} during the first 20 steps. The agent strategically balances the exploration of high-uncertainty areas with the exploitation of severe damage zones. Instead of wandering randomly, it systematically navigates from minor damage inland areas (I) toward the highly critical severe and destruction zones (levels III--IV) along the shoreline. By occasionally revisiting boundary regions to reduce overall variance, the agent efficiently reconstructs the damage distribution using far fewer measurements than grid-based or random schemes would require.

\begin{figure*}[!htbp]
  \centering
  \includegraphics[width=0.8\linewidth]{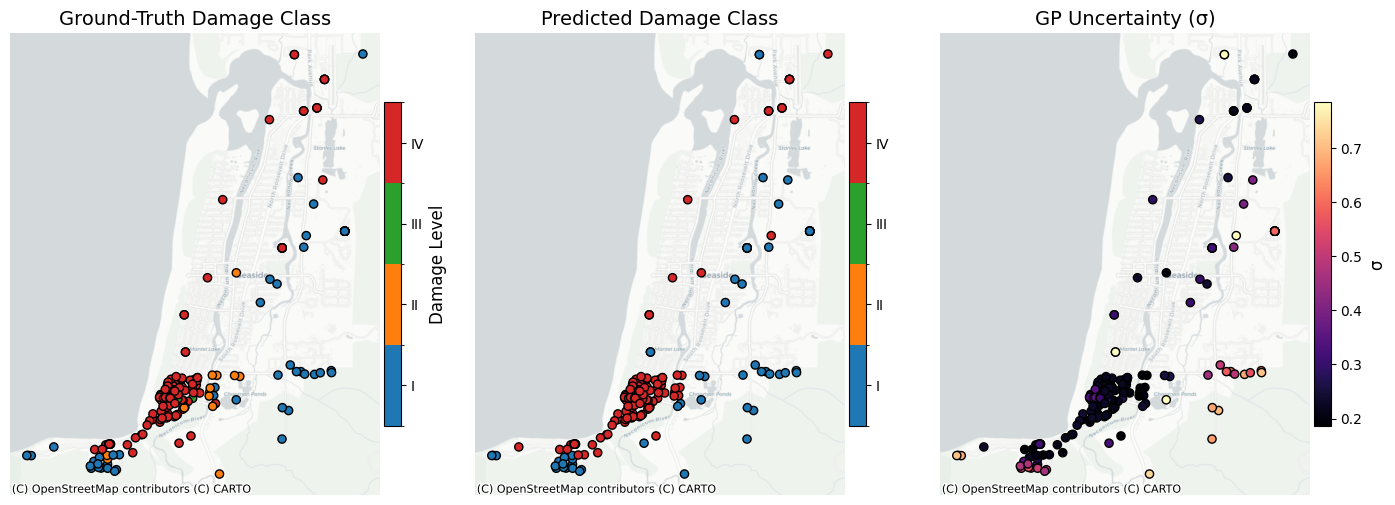}
  \caption{Spatial Comparison of Model Performance on Tsunami Damage Map. (left) Ground-truth building-level damage classes; (middle) Predicted classes from ODGP-CAF; (right) Posterior predictive uncertainty.}
  \label{fig:gpfinal}
\end{figure*}
Finally, Figure~\ref{fig:gpfinal} confirms the spatial effectiveness of the \textit{ODGP-CAF} model. The model accurately reconstructs the ground-truth damage pattern (left vs. middle), successfully identifying the critical level-IV destruction clusters. The posterior uncertainty map (right) corroborates the efficiency of the learned policy: the severely damaged coastal areas—which were prioritized and visited most frequently—exhibit the lowest variance ($\sigma \approx 0.2$), whereas sparsely sampled inland corridors retain higher uncertainty. This spatial alignment proves that the proposed cost-aware agent successfully prioritizes data collection in the most critical, high-risk regions.

\section{Conclusion.}
\label{sec:conclusion}
In this study, we proposed a cost-aware BO framework to predict the impacts of natural disasters through ordinal damage estimation. Our approach integrates a cost-sensitive acquisition function with level-set estimation, tailored for deep-kernel GP models in ordinal regression tasks. The framework was evaluated using a high-fidelity simulated tsunami scenario, demonstrating strong predictive performance and efficient sampling behavior. The proposed method is compared with several benchmarks that use an acquisition function that does not consider cost or randomly selects the next data point. In comparison to these benchmark methods, the proposed approach can recover the underlying damage map both accurately and with the lowest cost. Looking ahead, future work will explore more advanced model architectures by incorporating non-myopic BO strategies, aiming to further enhance long-term planning and decision making in adaptive damage assessment.

\bibliographystyle{siamplain}
\bibliography{example_references}
\end{document}